# M-FISH Karyotyping - A New Approach Based on Watershed Transform

Sreejini K S[1], Lijiya A[2] and V K Govindan[3]

Department of Computer Science and Engineering, National Institute of Technology, Calicut, Kerala, India

[1]`sreejini_k_s@yahoo.com`, [2]`lijiya@nitc.ac.in`, [3]`vkg@nitc.ac.in`

## ABSTRACT

*Karyotyping is a process in which chromosomes in a dividing cell are properly stained, identified and displayed in a standard format, which helps geneticist to study and diagnose genetic factors behind various genetic diseases and for studying cancer. M-FISH (Multiplex Fluorescent In-Situ Hybridization) provides color karyotyping. In this paper, an automated method for M-FISH chromosome segmentation based on watershed transform followed by naive Bayes classification of each region using the features, mean and standard deviation, is presented. Also, a post processing step is added to re-classify the small chromosome segments to the neighboring larger segment for reducing the chances of misclassification. The approach provided improved accuracy when compared to the pixel-by-pixel approach. The approach was tested on 40 images from the dataset and achieved an accuracy of 84.21 %.*

## 1. INTRODUCTION

In clinical and research cytogenetic studies, automated computerized systems for human chromosome analysis are very essential since a small deviation from the usual number of chromosomes will result in physical abnormalities. Chromosomes are structures located in nuclei of eukaryote cells that carry all the genetic instructions for making living organisms. Normal human metaphase spread contains 46 chromosomes, 22 pairs of autosomes and sex chromosomes (XY: Male, XX: Female). Chromosomes are present in every cell except red blood cells. Chromosome analysis is done on dividing cells in their metaphase stage (different phases of cell division: metaphase, anaphase, and telophase). During metaphase, chromosome can be stained to become visible and can be imaged by a microscope. Cells used for chromosome analysis are usually taken from amniotic fluid or from blood samples.

Karyotype is the tabular representation of human chromosomes in a cell. In this representation, the chromosomes are ordered by length from largest (chromosome 1) to smallest (chromosome 22 in humans), followed by sex chromosomes. Karyotypes are very useful for accurately diagnosing the genetic factors behind various diseases. Manual karyotyping is time-consuming, expensive and need well trained personnel. During the early period of chromosome analysis, researchers used grayscale images and features such as size, shape, centromere position and banding pattern for classification.

Since 1996, a staining method called M-FISH introduced by Speicher et al. [1] produces color images. This simplifies the karyotyping and detection of subtle chromosome aberrations. Images are captured with a fluorescent microscope with multiple optical filters. Combinatorial labelling of 5 fluorophores is used to assign a specific fluor combination to each of the chromosomes, so that each chromosome type can be visualized in a unique color. A sixth fluorophore, DAPI (4 in, 6-diamidino-2-phenylindole), is counterstained to all chromosomes. Figure 1 shows the five channel M-FISH image data of a VYSIS probe.

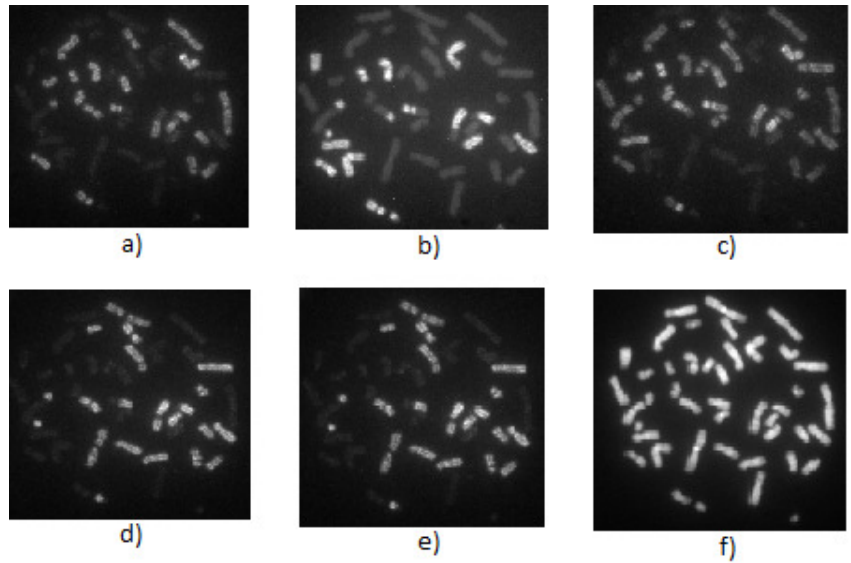

Figure 1. Five channel M-FISH image data. (a) Aqua fluor. (b) Red fluor
(c) Far red fluor. (d) Green fluor. (e) Gold fluor. (f) DAPI image.

This paper proposes an approach for M-FISH karyotyping based on naive Bayes classification of M-FISH image segments obtained by applying watershed segmentation. Here, classification is based on the features, mean and standard deviation, followed by a post-processing to reduce the misclassification.

The paper is organized as follows: Section 2 discusses some of the major existing work in the literature. Image segmentation and classification processes are given in Section 3. The comparative results obtained on standard database for the proposed approach and existing approaches are presented in Section 4, and the Section 5 concludes the paper.

## 2. LITERATURE SURVEY

Active research on karyotyping started since when the number of chromosome in human is found to be 46 in 1956. There are already a number of attempts proposed by various researchers to automate the process of karyotyping. We briefly review some of the major such works in this section:

The first M-FISH system developed by Speicher et al. [1] achieved semi-automated analysis of M-FISH image. In this approach, a mask is created to segment the chromosome from DAPI channel, and a threshold is applied to each pixel in the mask in order to detect the presence or absence of fluor in that pixel. The pixels are then classified by comparing its response with fluor labelling table. This unsupervised classification method, though simple requires some manual corrections of the segmentation map.

Presently used classification methods are based either on pixel-by-pixel or on region based classification algorithms. Pixel-by-pixel methods either classify each pixel of the M-FISH image or create a binary mask of the DAPI image using edge detection algorithms, and then classify each pixel of the mask. In region based methods, the regions obtained by decomposing the image are classified.

Automatic pixel-by-pixel classification approach proposed by Sampat et al. [2] modelled the karyotyping as a 25 class 6 feature pattern recognition problem and classified by using Bayes classifier. The 25 classes are the 24 chromosome types and the background, and the 6 features used are the gray scale values from 6 color channels. The classifier was trained, and tested only on a small set of non-overlapping images.

Another work of Sampat et al. [3] proposes supervised parametric and nonparametric classification techniques for pixel-by-pixel classification of M-FISH images. In supervised parametric technique, they modelled the problem as a 6-feature 25-class maximum likelihood pattern recognition task, and in the supervised non-parametric approach, they employed nearest neighbor and the k-nearest neighbor methods. Non-parametric approach performed better than the parametric approach. The highest classification accuracy was obtained with the k-nearest neighbor method and k = 7 is an optimal value for this classification task. The approaches do not handle overlapping images and they used only a small number of test images.

Unsupervised classification method based on fuzzy logic classification and a prior adjusted reclassification that corrects misclassifications is discussed in [4]. First, the separation of foreground and background is carried out by majority voting among k-means clustering, adaptive thresholding, LoG edge detection, and global thresholding methods. It requires spectral information, obtained from color table and then classifies the pixels by using fuzzy logic classifier and a prior adjusted reclassification was performed by adjusting the prior for each chromosome, so most likely class for each chromosome was found. It does not require training. High average accuracy is reported, however only a small number of test images were used.

Various pre-processing methods such as image registration, dimension reduction and background flattening are discussed in [5, 6]. Color compensation techniques    are discussed in [7]. The authors report that these techniques are useful for improving the accuracy of karyotyping.

Mohammed et al.  [8] presented an automated method for segmentation and classification of multispectral chromosome images. They used adaptive thresholding and valley searching for background cancellation. Discrete Wavelet Transform (DWT) is used to extract suitable number of features.  The approximate normalised DWT coefficients are used to reduce the size of the image to 325 X 260 pixels from 645 X 517 pixels.  Bayes decision theory is used to classify each pixel in the normalized approximation image. After classification, expands the size of image to its original, by adding zeros between each two neighboured pixels.  Majority filtering is used for removing the noise introduced during the expansion process. Each chromosome is segmented by collecting all the pixels belonging to this chromosome. The overlapping problem can be solved by use of the medial axis transform. High classification rate was reported.

Use of Gaussian mixture model (GMM) classifier for M-FISH images classification is presented in [9]. They modelled karyotyping as 26 class 6 feature pixel-by-pixel classification problem. The 26 classes are the 24 types of chromosomes, the background and the chromosome's overlap; the 6 features are the brightness of dyes at each pixel in six color channels. The overall classification accuracy achieved is reported as 89.18% and is found to be better than pixel-by-pixel method, but they used only a small number of images.

A region based watershed segmentation method applied to DAPI channel for multispectral chromosome image classification is presented in [10]. In this, marker controlled watershed transform is used to control over segmentation. A binary mask of the DAPI channel is computed in order to further reduce unwanted areas. Finally, a vector containing 5 features, each feature representing the average intensity value of each channel, is computed from each segmented area, and the vectors are classified using Bayes classifier. Good overall accuracy is reported. However, only a small number of non-overlapping testing images were used. This work was further extended with multichannel in [11]. They used gradient computed from all the channels. Classification is performed using region based Bayes classifier and the neighboring regions are then merged. This makes the detection of unhybridized regions simpler. Good overall accuracy is reported.

Support Vector Machines (SVM) classifier with multichannel watershed transform to perform M-FISH karyotyping was described in [12]. They constructed RB-SVM by using radial basis function as the kernel function. The method tested on images from normal cells and reported 10.16% increase in classification accuracy than Bayesian classification.

The work by Wang [13] deals with Fuzzy c-means clustering algorithm (FCM) based classification of M-FISH images. This uses 24 different cluster centers, which are formed from 24 classes of chromosomes and a pixel is assigned to each individual cluster according to its nearest distance to the center. Finding a cluster center is equivalent to minimizing the dissimilarity function; here, Euclidian distance is used as dissimilarity measure. The advantage of FCM is that it can locate centers more accurately because here the membership values are from 0 to 1. It works better than k-means clustering and Bayes classifier. Use of image normalization techniques such as image registration, dimension reduction and background subtraction are also used, leading to improvements in accuracy.

Later, Cao and Wang [14] presented Segmentation of M-FISH images for improved classification of chromosomes with an adaptive Fuzzy C-Means clustering. Adaptive FCM was done by incorporating a gain field which models and corrects intensity homogeneity and also regulates center of each intensity cluster. Intensity homogeneity is mainly caused by the image acquirement and uneven hybridization. It provides lowest segmentation and classification error and is better than FCM and AFCM.

Overlapping and touching chromosomes are still a problem in pixel-by-pixel classification. Many researchers have attempted to resolve this issue. Some of the important work in this category is [15, 16, 17]. In [15], minimum entropy is used as the main segmentation criterion to decompose overlapping and touching chromosome images. It also uses multi-spectral information in chromosome images. However, the computational time and complexity is very high; performance is very sensitive to its parameters and the approach is tested only on small number of images.

Extension of the above minimum entropy algorithm [15] is discussed in [16] which removes the pixel classification requirement. It works by estimating entropy from the raw data using differential entropy estimation technique, i.e., they used nearest neighbour estimation technique rather than calculating entropy from the classified pixels. The approach leads to computational complexity lower than that of minimum entropy approach. Still, the computational time required is unacceptably high, performance is very sensitive to its parameters and the algorithm is tested only on few images.

Another approach to resolve overlapping and touching of chromosomes is presented in [17]. The approach utilizes the geometry of a cluster, pixel classification results and chromosome sizes. First, the chromosomes are segmented from the background by majority voting among k-means clustering, adaptive thresholding, LoG edge detection, and global thresholding methods and then chromosome pixels are classified by using fuzzy logic classifier. A group of connected pixels is defined as cluster. Three sets of basic elements of cluster are cross shape cluster, T shape cluster and I shape cluster. For a given cluster, landmark on the boundary and skeleton are computed and then the cluster is decomposed in to multiple hypotheses, and the likelihood of each hypothesis is computed based on pixel classification results and chromosome size. Most likelihood hypothesis is chosen as the correct decomposition of that cluster. Good results are reported.

From the above review work on the various approaches suggested by the researchers, one can conclude that region based classification approaches are superior to pixel by pixel approach in terms of accuracy and computational time. Many researchers have tested their approaches only on small sets of selected images. For practical purposes, the systems must be capable of providing high accuracy on large data set. So, there is a need for research to improve the performance further on large data sets so that such automated systems for karyotyping can be acceptable for commercial purposes.

# 3. METHODS

The present work employs basically two major processing steps, segmentation and classification. Marker controlled watershed transform is used for segmentation and region based Bayes classification is used for classification.

## 3.1. CHROMOSOME SEGMENTATION

The separation of each chromosome from the metaphase image is the major operation carried out in this stage. Basic steps involved are: removal of cells from the DAPI images, gradient computation and minima selection, computation of watershed transformation and binary mask creation. These are briefly presented in the following subsections:

### 3.1.1. REMOVAL OF CELLS FROM THE DAPI IMAGE

Original DAPI chromosome image contains nuclei and debris along with chromosome. We must remove them based on the size and circularity before segmentation. Figure 2 shows the image before and after blob removal.

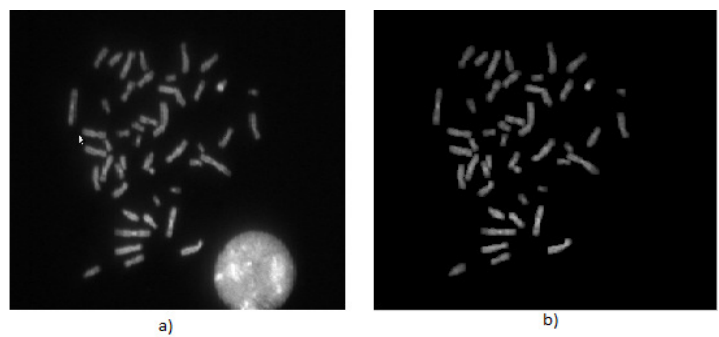

Figure 2. (a) Before blob removal (b) After blob removal

### 3.1.2. GRADIENT COMPUTATION AND MINIMA SELECTION

Gradient magnitude of the DAPI channel image after cell removal is computed. Sobel operators are used. Since watershed algorithm produces over segmentation, can be control by reduce the number of allowable minima in the gray scale.

### 3.1.3. COMPUTATION OF WATERSHED TRANSFORM

Watershed transform of the resulting image is computed which results in tessellation of the image in to different regions. The watershed transform has the advantage that the lines produced are always form closed and connected regions and these lines always correspond to obvious contours of objects which appear in image.

A gray scale image can be considered as a topographic surface, where height of each point is related to its gray level. If we punch a hole in each local minimum and immerse this surface in water, the regions in the image will start filling up with water. Immersion will starts from the points of minimum gray value. When water level in two or more adjacent basins will start merging, dams are built in order to prevent this merging. The flooding process will continue up to the stage at which only the top of dam is visible above the water line [18]. Watersheds are the lines dividing two catchment basins, each basins corresponds to each local minimum. Figure 3 shows the watershed lines superimposed on DAPI channel.

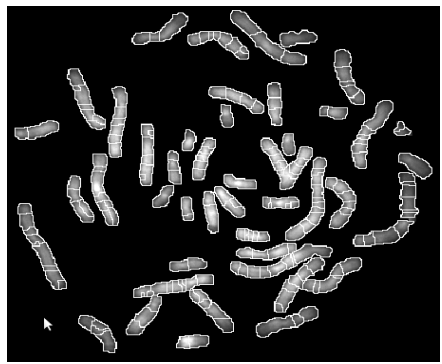

Figure 3. Watershed lines superimposed on DAPI channel

### 3.1.4. BINARY MASK CREATION

Binary mask is created from DAPI image after cell removal and superimposing watershed regions on it, in order to avoid the segmentation errors present due to unhybridization. Binary mask is created by Otsu's thresholding method [19]. Basic operation is, logical AND operation between the watershed lines and blob removed DAPI image.

## 3.2. FEATURE EXTRACTION AND CLASSIFICATION

### 3.2.1. FEATURE EXTRACTION

This stage extracts the features used for classification. Here, mean and standard deviation of each segmented area are the features used. Then the intensities of the pixels belonging to that region are replaced with mean intensity of that region for each segmented area.

### 3.2.2. CLASSIFICATION

The segments are classified using naive Bayes classifier. A naive Bayes classifier is a simple probabilistic classifier based on Bayes theorem with strong (naive) independence assumptions. Our goal is to classify the 46 chromosomes in to 24 chromosomes type (C = 24).

Let $x \in R^d$ denotes the feature vector computed from each segmented area; d = 5 X 2 = 10. Here, classification is done by using mean with standard deviation of the image under test. Let $P(c_i)$ denote the probability that a feature vector belongs to class $c_i$, where i varies from 1 to 24, and is called prior probability. Let p $(x \mid c_i)$ denotes the class conditional probability distribution function for a feature vector $x$ given that $x$ belongs to class $c_i$ and P $(c_i \mid x)$ be the posterior probability that the feature vector $x$ belongs to class $c_i$, given the feature vector $x$.

By using Bayes theorem,

$$P(c_i \mid x) = \frac{p(x \mid c_i)P(c_i)}{\sum_{i=1}^{24} p(x \mid c_i)P(c_i)} \tag{1}$$

Computed prior class probabilities from training samples are,

$$P(c_i) = \frac{(no.\ of\ regions\ belongs\ to\ class\ c_i)}{\sum_{k=1}^{24}(no.\ of\ regions\ belongs\ to\ class\ c_k)} \tag{2}$$

The general multivariate Gaussian density function [20] in d dimension is given by

$$p(x \mid c_i) = \frac{1}{(2\pi)^{d/2}|\Sigma_i|^{1/2}} exp(-\frac{1}{2}(x - \mu_i)^t \Sigma_i^{-1}(x - \mu_i)) \tag{3}$$

where $x$ is the d-dimensional feature vector from five channels and $\mu_i$ is the mean vector of each class $c_i$, $\sum_i$ is the d x d covariance matrix of the class $c_i$, and $|\sum_i|$ and $\sum_i^{-1}$ are the determinant and inverse. Also $(x - \mu_i)^t$ denotes the transpose of $(x - \mu_i)$.

For each class, we need to calculate $P(c_i \mid x)$, the class to which a feature $x$ belongs, is decided by Bayes decision rule.

$$Decide\ c_i, if\ P(c_i \mid x) > P(c_j \mid x), \forall j \neq i. \tag{4}$$

### 3.2.3. NEIGHBOR REGION MERGING

In this stage, for each region all the neighboring regions that share the same class are connected inorder to get meaningful class map. If regions are adjacent then those regions are connected in Region Adjacency Graph (RAG) and have a common boundary. Figure 4.a shows the original classmap and classmap obtained after neighbour region merging is shown Figure 4.b. Here each type of chromosome is colored with different color.

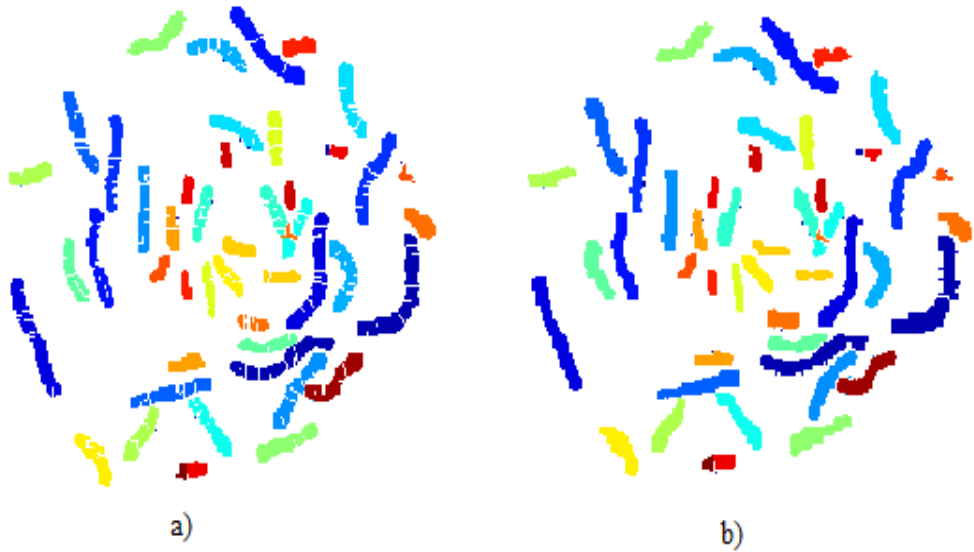

Figure4. (a) Before merging (b) After merging

### 3.2.4. POST-PROCESSING

It is observed that small segments are usually misclassified. To overcome this, small segments are reclassified to the most likely class of one of its neighbours by Bayes theorem, so that it becomes the same class as one of these neighbours.

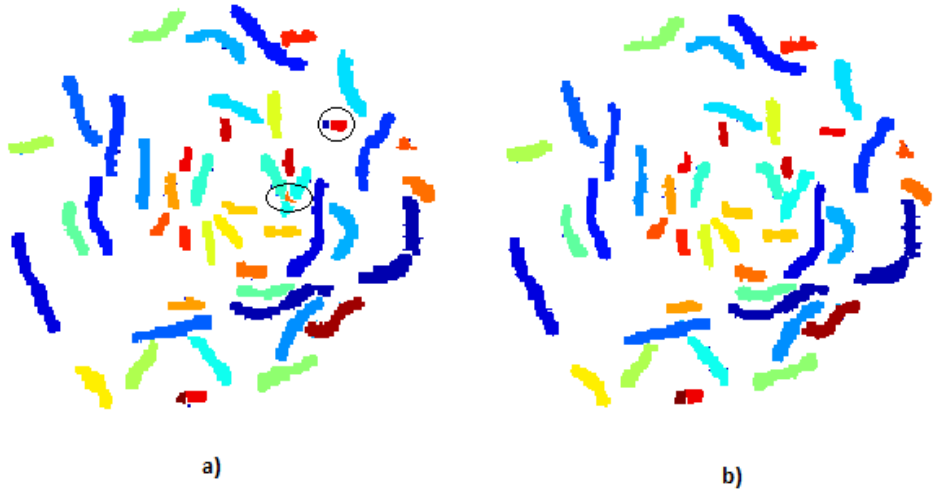

Figure5. (a) Before post-processing (b) After post-processing

Figure 5.a shows the results before post-processing. Circle denotes the misclassification that was corrected by the use of post-processing method as shown Figure 4.b.

# 4. RESULTS

## 4.1 M-FISH CHROMOSOME IMAGE DATABASE

Dataset [21] consist of 200 Multispectral images of size 517 X 645 pixels. 17 images are marked as extreme (EX), that are "difficult to karyotype". ASI, PSI, Vysis are the probes used. Each M-FISH image set consist of 5 monospectral images recorded at different wavelengths, DAPI and its "ground truth" image according to ISCN (International System for Human Cytogenetic Nomenclature) for each M-FISH image except for EX images. Ground truth image is labelled so that the gray level of each pixel represents its class number (chromosome type); background pixel values are zero; pixels in the overlapped regions values are 255. It is used to determine the accuracy of M-FISH images classification. But translocations are marked such that the full chromosome is labelled with the class which makes up the most of the chromosome. Images used for training and testing are taken from this dataset.

## 4.1 MINIMA SELECTION VALUE

In this method, the minima selection value is very important and it was found heuristically after several experiments and fixed to 5. As this value increases, area of each region is increased and total number of regions is decreased. Figure 6 shows the watershed segmentation of an M-FISH image with different minima selection values. White lines indicate the watershed lines, which are overlaid on M-FISH image.

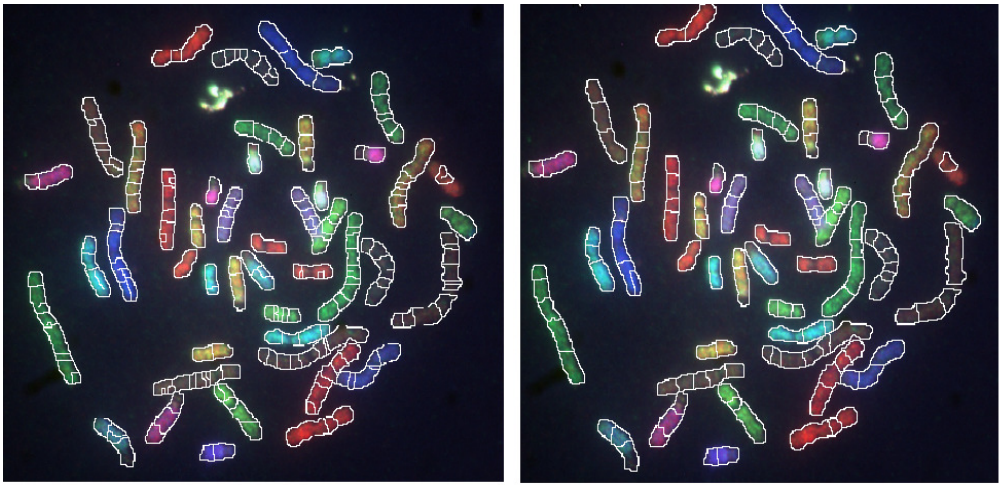

Figure 6: Watershed segmentation of an M-FISH image
(a) t = 5, #regions = 302 (b) t = 10, #regions = 194

## 4.1 SEGMENTATION AND CLASSIFICATION ACCURACY

To compare the performance of segmentation and classification, we need to define two figures of merits, namely, segmentation accuracy and classification accuracy [11]

Segmentation accuracy is defined as

$$Seg.\ Acc. = \frac{\#chromosome\ pixels\ correctly\ segmented}{\#total\ no\ of\ chromosome\ pixels} \qquad (5)$$

This method provides segmentation accuracy of 98.19% with standard deviation of 2.57%. Classification accuracy, is defined as

$$Class.\ Acc. = \frac{\#chromosome\ pixels\ correctly\ classified}{\#total\ no\ of\ chromosome\ pixels} \qquad (6)$$

Tables 1 show the comparison of classification accuracy obtained with various approaches, namely, pixel-by-pixel [2], mean only [10] and proposed approaches - mean & standard deviation, and mean & standard deviation with post-processing. Same training and testing images are used for all methods. In the present work, 40 images from the dataset are used for testing and the proposed approaches (mean & standard deviation, and mean & standard deviation with post-processing) give improved results compared to other methods [2, 10]. For all of the methods, classification accuracy can be further improved by proper pre-processing techniques [5 - 7]. The summary of the classification performance with 10 existing works is given in Table 2. Average classification accuracies of the proposed approach on best 5, 10, 15 and 40 images are also given in the Table 2. The results demonstrate that the performance of the proposed post-processing based approach is superior to most of the existing approaches.

*Table 1: Classification Accuracy*

| No. | Classification Accuracy of Various Approaches | | | |
|---|---|---|---|---|
| | | | Proposed | |
| | Pixel-by-pixel [2] | Mean [10] | Mean & Std. Dev. | Mean & Std. Dev. with Post-processing |
| 1 | 87.82 | 93.65 | 94.9 | 95.21 |
| 2 | 91.93 | 94.16 | 94.69 | 94.73 |
| 3 | 64.61 | 92.2 | 92.31 | 92.83 |
| 4 | 90 | 89.88 | 92.6 | 92.6 |
| 5 | 83.63 | 89.18 | 90.41 | 91.33 |
| 6 | 87.43 | 88.97 | 90.56 | 90.74 |
| 7 | 89.78 | 90.63 | 90.99 | 90.33 |
| 8 | 90.57 | 88.73 | 89.71 | 90.26 |
| 9 | 87.03 | 88.43 | 89.91 | 90.09 |
| 10 | 67.15 | 88.91 | 89.36 | 89.69 |
| 11 | 80.79 | 88.17 | 88.77 | 89.24 |
| 12 | 61.24 | 82.37 | 87.29 | 88.22 |
| 13 | 80 | 87.72 | 88.89 | 89.16 |
| 14 | 63.12 | 81.73 | 84.39 | 86.28 |
| 15 | 69.51 | 77.17 | 84.68 | 85.89 |
| 16 | 64.25 | 74.56 | 84.49 | 85.73 |
| 17 | 72.93 | 80.84 | 84.56 | 85.38 |
| 18 | 87.82 | 83.51 | 84.26 | 84.31 |
| 19 | 84.1 | 85.81 | 83.96 | 84.3 |
| 20 | 80.3 | 82.03 | 83.57 | 84.01 |
| 21 | 78.13 | 81.24 | 83.03 | 83.81 |
| 22 | 70.5 | 80.2 | 83.27 | 83.76 |

| | | | |
|---|---|---|---|
| 23 | 82.36 | 84.65 | 82.03 | 82.62 |
| 24 | 71.57 | 76.18 | 77.75 | 80.85 |
| 25 | 69.64 | 77.45 | 80.18 | 80.82 |
| 26 | 61.38 | 76.92 | 80.11 | 80.55 |
| 27 | 55.35 | 82.07 | 78.27 | 80.53 |
| 28 | 61.76 | 66.43 | 80.4 | 80.4 |
| 29 | 67.45 | 77.84 | 79.16 | 80.2 |
| 30 | 81.13 | 79.92 | 79.73 | 80.16 |
| 31 | 82.67 | 79.88 | 79.28 | 79.73 |
| 32 | 63.92 | 74.11 | 79.52 | 79.41 |
| 33 | 75.5 | 76.49 | 78.49 | 78.97 |
| 34 | 74.03 | 76.79 | 77.4 | 78.76 |
| 35 | 68.81 | 68.07 | 77.44 | 78.32 |
| 36 | 58.23 | 67.46 | 75.92 | 77.56 |
| 37 | 76.42 | 79.19 | 76.98 | 76.98 |
| 38 | 64.92 | 65.57 | 73.74 | 75.82 |
| 39 | 71.8 | 75.22 | 75.19 | 75.19 |
| 40 | 72.62 | 78.6 | 73.35 | 73.6 |
| **Avg** | **74.81** | **81.32** | **83.54** | **84.21** |

Table 2: Comparison of classification Accuracy of proposed and existing works

| Researcher | Average Chromosome Classification Accuracy | Number of Images Tested |
|---|---|---|
| Sampat et. al.[2] | 91.40% | n/a |
| Choi et.al.[22] | 85.90% | 5 |
| Wang [6] | 60.36% | 6 |
| Sampat et. al.[ 3] | 90.5% | 5 |
| Wang et. al [5] | 87.5% | 5 |
| Choi et. al.[ 4] | 89.08% | 9 |
| Schwartzkopf [24] | 68.00% | 183 |
| Karvelis P S [10] | 89.53% | 15 |
| Fazel [9] | 89.17% | 6 |
| **Proposed** | **84.21%** | **40** |
| | **93.34%** | **Best 5** |
| | **91.70%** | **Best 10** |
| | **90.44%** | **Best 15** |

### 4.1. CLASSIFICATION MAP

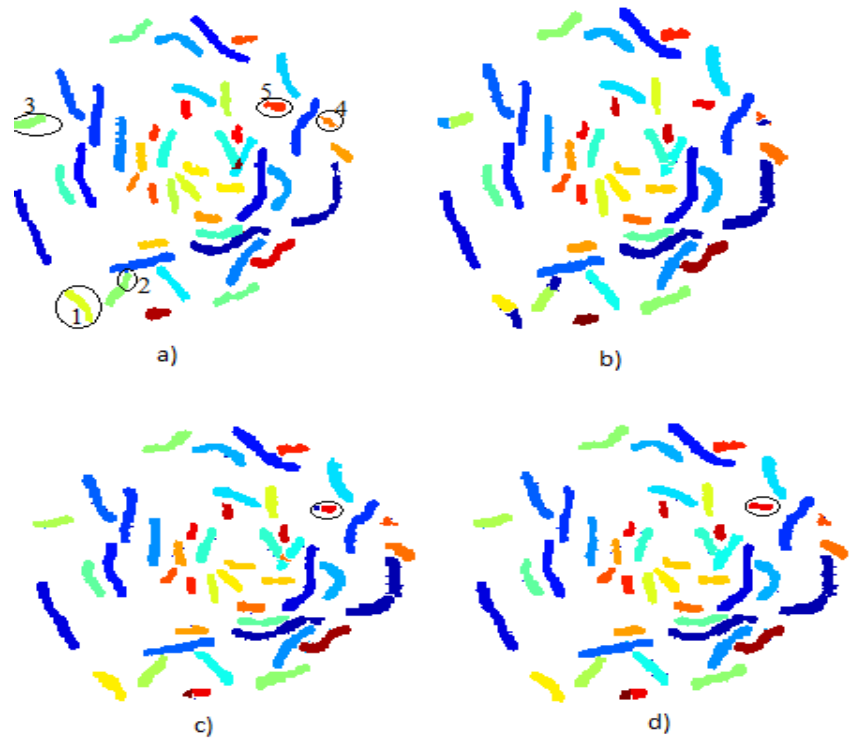

Figure 7:  Classmap obtained for various approaches (a) Ground Truth (b) Mean only method (c) Mean and standard deviation method (d) Proposed method

The classification map obtained for different approaches are shown in Figure 7.  Actual ground truth is shown in Figure 7.a. The chromosomes pixels marked in circle 1, 2, 3, 4 and 5 are same chromosome type pixels. Chromosomes pixels marked in 1, 2, 3 and 4 are misclassified by mean only method and as shown in Figure 7.b. The chromosome pixels in 5 are misclassified, as shown in Figure 7.c, when classifying with mean and standard deviation approach. This is correctly classified, as shown in Figure 7.d, when post-processing technique was applied.

## 5. CONCLUSION

An automated method for M-FISH karyotyping employing watershed segmentation followed by naïve Bayes classifier is presented. Mean and standard deviation are the features used for classification. Improved accuracy is obtained by adding a post-processing method that re-classifies the small segments to neighboring segments based on Bayes theorem. This method works for all probes and the results are better than pixel-by-pixel classification, which always produces noisy results. As the classification is done on the watershed regions, the computational time needed is also much less than the pixels by pixel approach. Use of pre-processing techniques [5, 6] and manually corrected ground truth [22] will further improve classification accuracy. Future work is to extend this method for multichannel watershed and to test on much larger data set.

# REFERENCES

[1] Speicher M.R., S. G. Ballard, and D. C. Ward, "Karyotyping human chromosomes by combinatorial multi-fluor FISH," *Nat. Genet.*, vol. 12, pp. 368 – 375, 1996.

[2] Sampat M. P., A. C. Bovik, J. K. Aggarwal, and K. R. Castleman, "Pixel-by-Pixel Classification of MFISH Images," Proc. 24[th] IEEE Ann. Intern. Conf. (EMBS), Houston, TX, 2002, pp. 999-1000.

[3] Sampat M. P., A. C. Bovik, J. K. Aggarwal, and K. R. Castleman, "Supervised parametric and non-parametric classification of chromosome images," *Pattern Recognit.*, vol. 38, pp. 1209–1223, Aug. 2005.

[4] Choi H., K. R. Castleman, and A. C. Bovik, "Segmentation and fuzzy-logic classification of M-FISH chromosome images," in *Proc. IEEE Int. Conf. Image Process. (ICIP 2006)*, Atlanta, GA, pp. 69–72.

[5] Wang. Y. and K. R. Castleman, "Normalization of multicolor fluorescence in situ hybridization (M-FISH) images for improving color karyotyping," *Cytometry*, vol. 64, pp. 101–109, Apr. 2005.

[6] Wang Y., "M-FISH image registration and classification," in *Proc. IEEE Int. Symp. Biomed. Imag.: Nano to Macro*, Apr. 2004, vol. 1, pp. 57–60.

[7] Choi H., K. R. Castleman, and A. C. Bovik, "Color Compensation of Multicolor FISH Images", In IEEE Transaction on Medical Imaging, volume 28, January 2009.

[8] Mohammed Alhanjouri, Fatma E. Z. Abou - Chadi, and Nadder Hamdy, "Segmentation and Classification of Multispectral Chromosome Images," Twenty Second National Radio Science Conference, March 15-17, Cairo-Egypt, 2005.

[9] Wang Y., Fazel A., and Derakhshani R, "Classification of multicolor fluorescence in situ hybridization images using gaussian mixture models," Proceedings of ANNIE Conference, 2006.

[10] Karvelis P. S., D. I. Fotiadis, M. Syrrou, and I. Georgiou, "A watershed based segmentation method for multispectral chromosome images classification," in *Proc. 28th IEEE Ann. Intern. Conf. (EMBS)*, New York, 2006, pp. 3009–3012.

[11] Karvelis P. S., A. T. Tzallas, D. I. Fotiadis, and I. Georgiou, "A multichannel watershed-based segmentation method for multispectral chromosome classification," IEEE Trans. on Med. Imag., vol. 27, no. 5, pp. 697-708, 2008.

[12] Georgiou. I., P. Sakaloglou, P. S. Karvelis and D. I. Fotiadis, "Enhancement of the Classification of Multichannel Chromosome Images using Support Vector Machines," 31[st] Annual International Conference of the IEEE EMBS, Minneapolis, USA, September 2-6, pp- 3601-3604, 2009.

[13] Wang Y., "Classification of M-FISH Images using Fuzzy C-means Clustering Algorithm and Normalization Approaches," *38[th] Asilomar Conference on Signals, Systems and Computers*, vol. 1, Issue 7-10, Nov, pp. 41 – 44, 2004.

[14] Cao H. and Y. Wang, "Segmentation of M-FISH images for improved classification of chromosomes with an adaptive Fuzzy C-Means Clustering Algorithm," In IEEE international symposium on biomedical imaging: from nano to macro, pp 1442–1445, 2011.

[15] Schwartzkopf W. C., B. L. Evans, and A. C. Bovik, "Minimum Entropy Segmentation Applied to Multi-Spectral Chromosome Images," Proc. IEEE Int. Conf. on Image Processing, Thessaloniki, Greece, vol. II, pp. 865-868, Oct. 7-10, 2001.

[16] Schwartzkopf W. C., B. L. Evans, and A. C. Bovik, "Entropy Estimation for Segmentation of Multi-Spectral Chromosome Images", IEEE Southwest Symposium on Image Analysis and Interpretation, April 7-9, pp.234-238, 2002.

[17] Choi H., A. C. Bovik, and K. R. Castleman, "Maximum-likelihood decomposition of overlapping and touching M-FISH chromosomes using geometry, size and color information," In Proceedings of the 28[th] IEEE EMBS annual international conference, 2006, pp 3130–3133, 2006.

[18] Vincent L., and P. Soille, "Watersheds in Digital Spaces: An Efficient Algorithm Based on Immersion Simulations," IEEE Transaction on Pattern Analysis and Machine Intelligence, 13, No.6, pp. 583–598, June 1991.

[19] Otsu N., "A threshold selection method from gray-level histograms," IEEE Trans on Syst., Man, Cybern., vol. 9, no. 1, pp. 62–66,Jan. 1979.

[20] Duda. R. O., P. E. Hart, and D. G. Stork, "Pattern Classification," San Diego: Harcourt Brace Jovanovich, Second ed., November 2000.

[21] http://dip4fish.blogspot.com/2011/11/mfish-dataset-available.html

[22] Hyo Hun Choi, "Automatic Segmentation and Classification of Multiplex-Fluorescence In-Situ Hybridization Chromosome Images", Phd Dissertation, The University of Texas at Austin, 2006.

[23] Choi H., K. R. Castleman, and A. C. Bovik, "Joint segmentation and classification of M-FISH chromosome images," in *Proc. 26th IEEE Annu. Int. Conf (EMBS)*, San Francisco, CA, 2004, vol. 1, pp. 1636–1639.

[24] Schwartzkopf W. C., A. C. Bovik, and B. L. Evans, "Maximum-likelihood techniques for joint segmentation-classification of multispectral chromosome images," IEEE Trans. Med. Imag., vol. 24, pp. 1593-1610, Dec.2005.